\documentclass{article}
\pdfoutput=1
\usepackage{spconf,amsmath,graphicx}
\usepackage{cite}

\newcommand{\comment}[1]{}

\title{Signal Clustering with\\Class-independent Segmentation}
\name
  {Stefano Gasperini$^{1,2}$ \quad Magdalini Paschali$^1$ \quad Carsten Hopke$^2$ \quad David Wittmann$^2$ \quad Nassir Navab$^{1,3}$}
\address{$^1$ Computer Aided Medical Procedures, Technical University of Munich, Germany\\
	$^2$ Airbus Defence and Space GmbH, Manching, Germany\\
	$^3$ Computer Aided Medical Procedures, Johns Hopkins University, Baltimore, USA}

\begin{document}
%
\maketitle
%

\label{parts:abstract}

\begin{abstract}

Radar signals have been dramatically increasing in complexity, limiting the source separation ability of traditional approaches. In this paper we propose a Deep Learning-based clustering method, which encodes concurrent signals into images, and, for the first time, tackles clustering with image segmentation. Novel loss functions are introduced to optimize a Neural Network to separate the input pulses into pure and non-fragmented clusters. Outperforming a variety of baselines, the proposed approach is capable of clustering inputs directly with a Neural Network, in an end-to-end fashion.

\end{abstract}
%

\section{Introduction}
\label{sec:intro}


Radars are used for situation awareness in a variety of applications, from weather forecasting to adaptive cruise control. Aircrafts utilize them also as a sensor, inspecting the received signals to get insights about their surroundings.
Towards this end, signals are initially isolated and then compared against prior knowledge for identification. In this paper, we focus on separating simultaneous and aligned signals by source.

Radar signals are composed of pulses, described by time of arrival (TOA), radio frequency (RF), pulse width (PW), amplitude (AM)  etc.~\cite{scheer2013radar_principles}. 
This task is challenging, especially because the separation is done at pulse level, based on a few parameters, that can be shared across multiple sources.


Furthermore, due to the rapid progress of antenna technologies and electronics~\cite{scheer2013radar_principles}, traditional methods cannot cope anymore with the increased complexity of incoming signals. Based on statistics and pattern matching, they struggle even with a few concurrent inputs~\cite{manickchand2017comparative}. To overcome this issue, the task has been addressed with data analysis clustering methods, among which DBSCAN reached superior performance and wide applicability~\cite{wang2013dbscan_radar, saab2019thesis_dbscan}.


Over the past years, Deep Learning (DL) has achieved state-of-the-art results in a plethora of tasks, including clustering~\cite{tum2018clustering}.
In the imaging domain, Neural Networks (NNs) are trained to map the inputs to a clustering-friendly representation~\cite{tum2018clustering, xie2016dec, huang2014den, ghasedi2017depict}. Afterwards, these newly extracted features are clustered with a commonly used data analysis method~\cite{tum2018clustering, min2018survey}, such as K-means~\cite{yang2017towards} or Hierarchical agglomerative clustering~\cite{yang2016jule}.
The same idea has been applied to the speaker separation task, using K-means~\cite{hershey2016deep}.
Pairing a NN with traditional clustering methods leads to improved performance, but inherits shortcomings from both approaches, such as requiring a predefined number of clusters and a sufficiently large dataset for training a NN.
Additionally, these methods would cluster pulses based on the similarity of their parameters, missing the aim of our task, which is grouping them by source.
Furthermore, mapping separable features requires large input dimensions, while our pulses have only a few parameters.

In radar signal processing, NNs have been mostly applied to address classification and identification problems~\cite{cain2018convolutional, matuszewski2018radar, li2016radar}. Recently, Recurrent NNs (RNNs) have been deployed to cluster pulses~\cite{liu2018classification}. Each RNN was trained to identify a specific signal and checked at every possible starting pulse of a sequence. Tested on 5 simple signals~\cite{liu2018classification}, this method would not be feasible in real world applications with thousands of emitters, each capable of producing multiple signals. It would require long processing times and could not separate those signals for which an RNN has not been trained.

\begin{figure}[t]
\centering
  \includegraphics[width=0.41\textwidth]{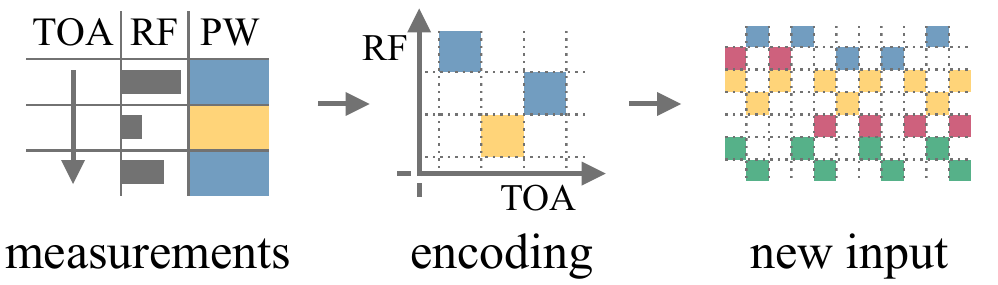}
   \caption{Domain change from signal to image processing.}
   \label{fig:encode}
\end{figure}



The contribution of this paper can be summarized as follows: 1) We propose
an original NN-based clustering method and we evaluate it on the challenging task of aligned radar pulse deinterleaving; 2) We introduce novel loss functions, aimed at delivering pure and non-fragmented clusters, while performing class-independent segmentation; 3) To the best of our knowledge, for the first time, we address clustering as an image segmentation problem.


\section{Method}

The idea is changing the problem domain towards imaging, to benefit from the extensive DL research done in the field.
Image segmentation is the task of separating pixels into regions according to some given criteria, alike clustering, which groups elements based on similarity.
Towards this end, we aim at exploiting image segmentation techniques to perform clustering and tackle the radar pulses deinterleaving problem.

Our proposed method enables a NN to cluster directly input elements through image segmentation. In this paper, we apply it on radar signals deinterleaving. Towards this end, we first encode the signals into spectrogram-inspired segmentable images (Section~\ref{domain_change}). These are then forwarded to a U-Net~\cite{ronneberger2015unet} trained for image segmentation to group the inputs. The NN is optimized with novel loss functions derived from a newly adjusted confusion matrix, that we named soft confusion matrix (Section~\ref{losses}). The objective functions aim at improving clustering performance indicators such as purity and fragmentation~\cite{thompson2006measures} (Section~\ref{conf_mat_metrics}).

\subsection{From Signals to Segmentable Images}
\label{domain_change}
To cross from the radar to the imaging domain, we need to encode the signals into images.
Spectrograms are graphical representations of signals, showing RF over time.
For radars, keeping the PW resolution within reasonable images of 512x512 pixels, would visualize too few pulses per signal, if any, complicating the clustering task.

\noindent \textbf{Signals encoding.} Towards this end, our representation is inspired by spectrograms, as we encode concurrent signals into an image, using it as a RF-TOA coordinate system.
We keep a time resolution of 5 $\mu s$, which allows for the encoding of a reasonable amount of pulses within each 512x512 image.
We indicate the pulses by marking in the RF-TOA grid those $(rf, toa)$ cells corresponding to incoming pulses at time $toa$, parametrized with an RF value of $rf$.
Pixels have gray-scale values representing PW and AM in two separate channels.
The encoding is schematized for PW in Fig.~\ref{fig:encode}, the same is done for AM.
We maximize the resolution by scaling each parameter individually to cover all its available range: [0, 512] for RF, and (0, 1] for PW and AM. This further diversifies the values, easing the separation task.
Moreover, we improve readability, by extending the values across the neighboring 3x3 pixels.
Overall, with this transformation no major loss of information occurs, and discretization can help against noise.

Our problem differs substantially from traditional image segmentation, since instead of identifying a preset amount of classes, we deal with an unknown and variable number of clusters.
Nevertheless, we can retrieve the cluster assignments from a segmented image, by remembering the pulses location from the encoding step, filtering the background and applying majority voting within each 3x3 region.

\subsection{Confusion Matrix-based Metrics}
\label{conf_mat_metrics}

\begin{figure}[t]
\centering
  \includegraphics[width=0.37\textwidth]{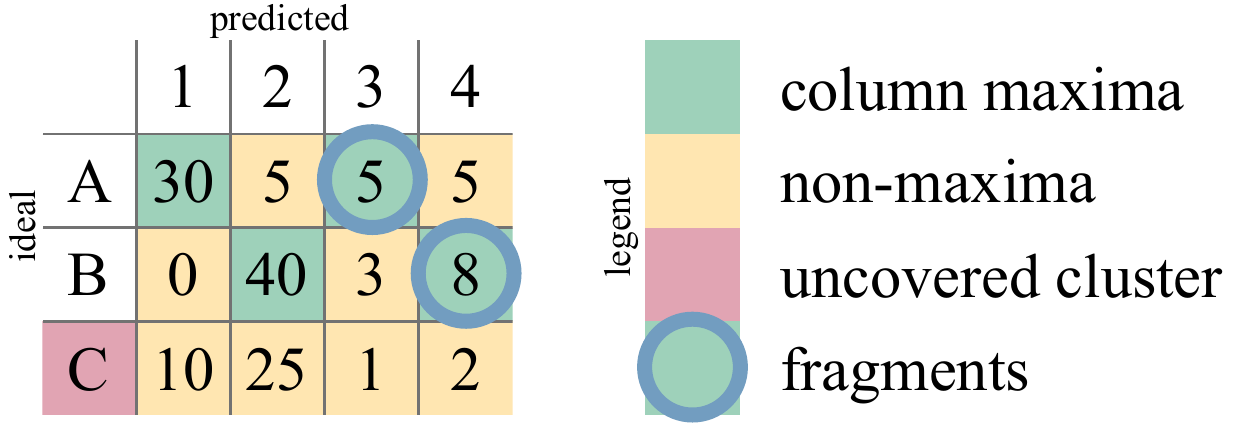}
   \caption{Confusion matrix and properties for evaluation.}
   \label{fig:conf_mat_legend}
\end{figure}

We choose a U-Net architecture~\cite{ronneberger2015unet}, due to its state-of-the-art performance in a variety of image segmentation tasks. While typically U-Nets are trained with pixel-wise cross-entropy or Dice loss, these functions are designed for a fixed amount of classes and are not suitable for clustering.

Instead, to address clustering problems, we derive new loss functions, that maximize performance metrics.
Such indicators can be formalized with the help of a confusion matrix~\cite{thompson2006measures}, like the one showed in Fig.~\ref{fig:conf_mat_legend}. The matrix has the predictions along the horizontal axis, with the cells showing the elements count. With $C_i^j$ we refer to the $(i, j)$ cell of the confusion matrix $C$, sized $G \times P$; with $C_i$ and $C^j$ to its $i$-th row and $j$-th column respectively.

\noindent \textbf{Cluster Purity.} The first step is identifying the column maxima $M(C) := \left\{ \max \left( C ^ { j } \right) \forall~j \right\}$, shown green in Fig.~\ref{fig:conf_mat_legend}. This matches each predicted cluster with a ground truth. In the example, what was predicted as cluster "4" corresponds to "B".
Cluster purity is defined as $cp = \sum {M(C)} / \sum {C}$: it assesses how many elements define the identity of each predicted cluster.
A system would score perfect purity predicting a different cluster for each input element, while performing rather poor.

\noindent \textbf{Fragmentation Ratio.} To address these cases, we additionally consider fragmentation ratio. It compares the number of fragments, marked with a circle in Fig.~\ref{fig:conf_mat_legend}, with the amount of predicted clusters $P$. It is defined as $cfr = \Sigma _i \max \left( 0, ~ \left| C_i \cap M(C) \right| -1 \right) / P$. Fragments occur when a row contains more than one column maxima.


\subsection{Overcoming Non-differentiability}
\label{losses}

The two metrics are non-differentiable: they are extracted from a confusion matrix, which is based on an argmax function over the assignment probabilities. To circumvent this issue, we take a step back and compute the losses directly from the softmax probabilities along the predicted clusters.

\begin{figure}[t]
\centering
  \includegraphics[width=0.41\textwidth]{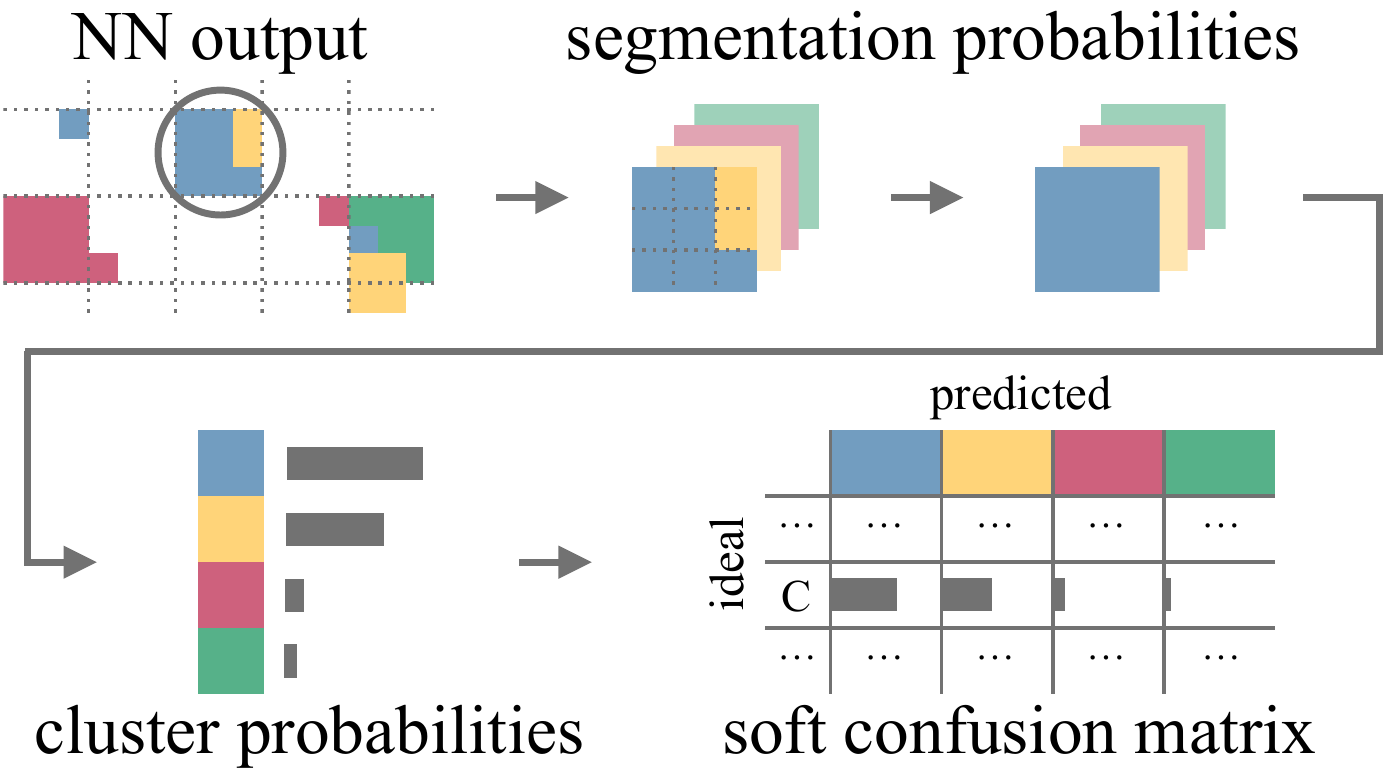}
   \caption{Construction of the soft confusion matrix.}
   \label{fig:soft_mat_constr}
\end{figure}

\noindent \textbf{Soft Confusion Matrix.} Towards this end, we create a new matrix, that we name soft confusion matrix (SCM). Its construction process is shown in Fig.~\ref{fig:soft_mat_constr}. Instead of reporting the discrete argmax assignments, we build the SCM with the softmax probabilities for each input element. We report the averages within its 3x3 pixels along its corresponding SCM row.
After repeating this for all input elements, we complete the SCM by summing the probabilities accumulated within each cell.
We construct the SCM using only differentiable operations, so we can compute our losses with its values.
Following the notation of Section~\ref{conf_mat_metrics}, $S$ is the SCM, sized $G \times N$, with $N$ being the amount of output channels of the NN.

\noindent \textbf{Soft Cluster Impurity Loss.} From the cluster purity metric (Section~\ref{conf_mat_metrics}), we derive its differentiable complement from the SCM. This is the cluster impurity loss, which maximizes the column maxima $M(S)$. It is defined as $sci\_l$ in Eq.~\ref{eq:cil}.

\begin{equation}
\begin{matrix}
    sci\_l = \frac{\sum Q(S)}{\sum S}\\
    \scriptstyle Q(S):= \scriptstyle \left\{ S_i^j \right\} : \: S_i^j  \notin M^j(S) \:\: \forall i,j
    \label{eq:cil}
\end{matrix}
\end{equation}

\noindent \textbf{Soft Cluster Fragmentation Loss.} Analogously, soft fragmentation loss $scf\_l$ corresponds to the fragmentation metric (Section~\ref{conf_mat_metrics}), which is computed with non-differentiable counting functions. We make it suitable for training, by applying the trick in Eq.~\ref{eq:cfl}, using only differentiable operations.

\begin{equation}
\begin{matrix}
    scf\_l = \frac{\sum F(S) \, \oslash \, F(S)}{N}\\
    \scriptstyle F(S) \, := \, \scriptstyle \left\{ S_i^j \right\} : \: S_i^j  \in R_i(S) \; \land \; S_i^j \, \neq \, \scriptstyle \max \left( R_i (S) \right) \:\: \forall i,j \\
    \scriptstyle R_i(S) \, := \, \scriptstyle S_i \; \cap \; M(S)
    \label{eq:cfl}
\end{matrix}
\end{equation}

With reference to Eq.~\ref{eq:cfl}: $F(S)$ is the set containing the SCM cells responsible for fragments, which are penalized by this loss; $R_i(S)$ is the set of SCM column maxima within the $i$-th row; $\oslash$ is the Hadamard element-wise division. Therefore $F(S) \oslash F(S)$ is a unit vector with a length corresponding to the amount of fragments, and the $scf\_l$ numerator is the differentiable equivalent of counting them.

Additionally, at each training iteration we randomly swap the cluster target values. With this strategy, we ensure the NN focuses on grouping, rather than learning specific associations. This way we also achieve class-independence.

The key of our method is the combination of the two aforementioned novel loss functions, which enables predicting the clusters directly, and addressing the clustering problem as an image segmentation task.

\section{Experimental Setup}

\noindent \textbf{Dataset.} Since there was no suitable public dataset available, we created one with the help of a domain expert. 140 realistic signals were defined and generated with an RF software simulator. Training and test sets were designed separately,
increasing the complexity: 75 signals are utilized for training, 65 for testing.
To create the dataset, we randomly made concurrent signals from different sources.
Since a good clustering method should deal with a variable amount of clusters, random 2.56 $ms$ portions of a signal appear alone, as well as with 1 to 10 other signals a variable amount of times. Concurrent signals are at most 11, all aligned, coming from the same direction, to be distinguished through TOA, RF, PW and AM. Overall, 1200 combinations are generated from the training signals and 300 from the testing, with the amount of pulses varying significantly throughout the dataset (min: 6, max: 1032, average: 467).

\begin{figure}[t]
\centering
  \includegraphics[width=0.47\textwidth]{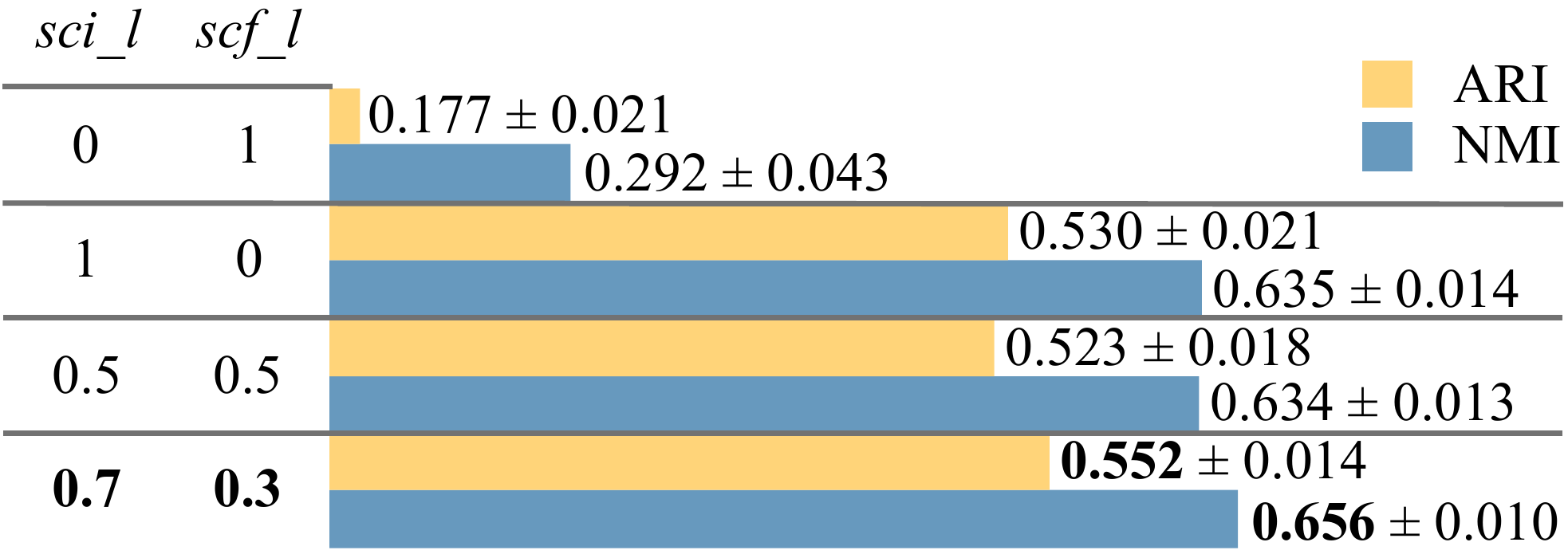}
   \caption{Ablative testing. Different configurations of the loss functions, indicated on the left columns, are evaluated with ARI and NMI. The last row shows the proposed method.}
   \label{fig:ablative}
\end{figure}

\noindent \textbf{Architecture.} A U-Net~\cite{ronneberger2015unet} was used for all experiments. To increase its robustness against overfitting and improve its suitability for embedded devices, we reduced its size: feature map channels start at 8 and reach 64 in the bottleneck layer. Since segmentation requires fixed amounts of classes, we set the concurrent predictable clusters to $N=15$, following~\cite{lukic2016speaker}.

\noindent \textbf{Model Training.} Across all experiments, hyperparameters remained constant, to ensure comparability. The U-Net was trained for 300 epochs with Adam optimizer, a learning rate of $10^{-5}$ and unit batch size. We optimized for the novel losses described in Section~\ref{losses}. We implemented our method in PyTorch and trained our models on an NVIDIA Titan Xp GPU.

\begin{figure*}[ht]
\centering
  \includegraphics[width=0.96\textwidth]{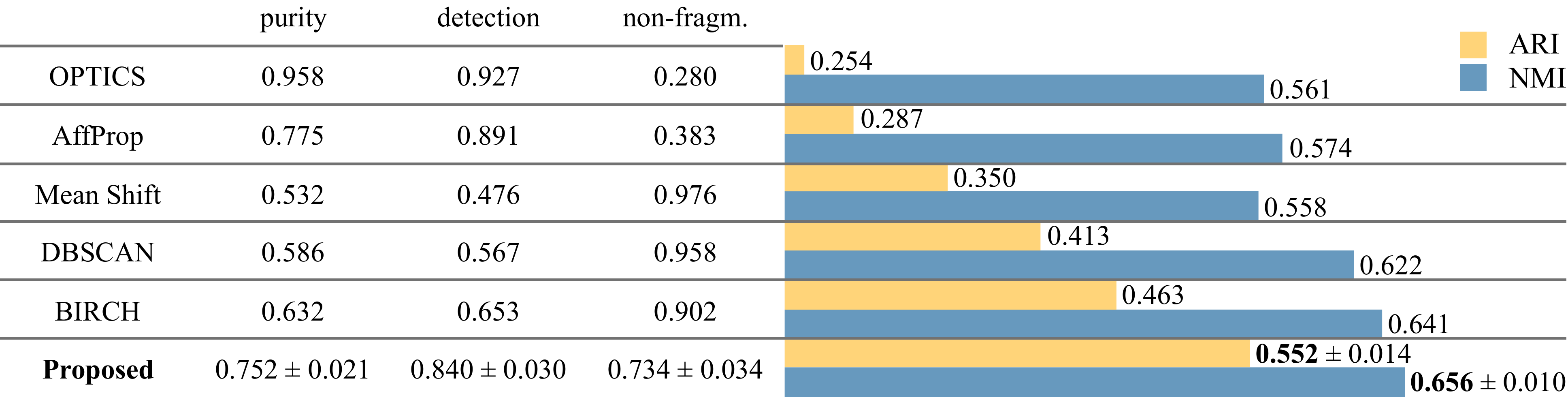}
   \caption{Comparison with baseline methods across five performance metrics.}
   \label{fig:baseline}
\end{figure*}

\noindent \textbf{Evaluation Metrics.} The evaluation is based on the criteria presented in Section~\ref{conf_mat_metrics}: cluster purity ($cp$) and (non-) fragmentation ratio ($cnfr$). On top of pure and non-fragmented clusters, we want to ensure that each ground truth has a corresponding predicted cluster. Specifically, a cluster remains uncovered (marked red in Fig.~\ref{fig:conf_mat_legend}) when its confusion matrix row contains no column maxima $M(C)$. This is measured by the detection ratio, which is defined as $cdr = 1- \left| U \right| / G$, with $U\!:=\!\left\{i\right\} \!\!: C_i \cap M(C) = \emptyset$.
These three metrics should be considered combined, since they provide a good assessment of the clustering performance from different important aspects. Additionally, we deployed standard clustering evaluation metrics, namely Adjusted Rand Index (ARI) and Normalized Mutual Information (NMI). We repeated all experiments five times, reporting mean and standard deviation.

\noindent \textbf{Ablative Testing.} In order to showcase the effectiveness of the main components of our method, we performed ablative testing. We evaluated the impact on the clustering performance of different loss function configurations.

\noindent \textbf{Baseline Comparison.} Furthermore, we compared our method against various clustering approaches. Despite being popular and effective in other domains, several methods, such as K-means, are not applicable to the task at hand, since they require a predefined number of clusters, which in our case is unknown and varies over time. We deployed five suitable baseline methods, namely DBSCAN, OPTICS, BIRCH, Affinity propagation and Mean shift~\cite{xu2005survey, xu2015comprehensive}.


\section{Results and Discussion}

\noindent \textbf{Ablative Testing.} Fig.~\ref{fig:ablative} showcases that combining the losses can improve the results regarding both ARI and NMI.
The purity loss achieves satisfying results when deployed by itself too, since its optimization acts on every cell of the confusion matrix (Section~\ref{conf_mat_metrics}), trying to improve all predictions. On the other hand, the fragmentation loss focuses only on a few cells, rendering it unable to deliver good overall performance by itself. However, optimizing them jointly in a weighted fashion of 0.7 for purity and 0.3 for fragmentation delivers higher ARI and NMI. The fragmentation loss is important: it contributes by penalizing the prediction of unnecessary clusters.

\noindent \textbf{Comparative Methods.}
Fig.~\ref{fig:baseline} highlights the superior performance of our method compared to the other clustering approaches. Most baseline methods suffer from either over predicting the amount of clusters (high $cp$, low $cnfr$), or underestimating it (low $cp$, high $cnfr$). BIRCH stands out among the other baseline approaches. Nevertheless, our method outperforms BIRCH as well with an improvement of 1.5\% for NMI and 8.9\% for ARI. This is accounted to the learning capacity of our method, which is trained to improve both $cp$ and $cnfr$ jointly. Repeating the experiments with the baseline approaches leads to the same clusters. Our method, despite relying on a random initialization of its model weights, is able to deliver consistent results with a small standard deviation.

\noindent \textbf{Metrics Trade-off.}
As can be seen in Fig.~\ref{fig:baseline}, there is often a trade-off between purity and non-fragmentation ratio scores. The two metrics are evaluating the performance from different perspectives and it is challenging to achieve high scores for both, especially in a task such as radar pulse deinterleaving. Instead, getting a perfect score for only one of the two is trivial, although decreasing the other metric substantially. Such unbalanced performances can be achieved by predicting a single cluster for maximum $cnfr$, or one per input for $cp$.
The strength of the proposed method lies in its ability to optimize both metrics simultaneously, improving two complementary aspects.

\noindent \textbf{Cross Domain Applicability.} The results achieved show good potential for a previously unexplored way of solving clustering problems. Despite being applied on the pulses deinterleaving task, the method can be extended to other domains: the novel soft confusion matrix-based loss functions could pave the way towards new NN-based clustering approaches that could operate in an end-to-end fashion, without requiring conventional data analysis methods. This would alleviate shortcomings of traditional clustering approaches, such as dealing with a predefined number of clusters, or dependencies to cluster shapes and point densities.



\section{Conclusion and Future Work}

In this paper we proposed a new DL-based clustering method, which we applied on the challenging task of aligned radar pulse deinterleaving. For the first time, clustering was targeted as an image segmentation problem. We changed domain by encoding the concurrent signals into segmentable images. The NN was trained to predict the clusters directly in an end-to-end fashion, aiming at pure and non-fragmented clusters, thanks to new loss functions derived from a novel probability-based confusion matrix. 

Furthermore, we plan on scheduling different loss weightings over training and exploring other input representations, suitable for a even larger amount of simultaneous and aligned signals. Moreover, we will adapt the proposed method and losses to be suitable for different domains, such as image clustering, comparing ours against other DL-based approaches.


\bibliographystyle{IEEEbib}
\bibliography{refs}

\end{document}